\documentclass[conference,twocolumn]{IEEEtran} 
\pdfoutput=1
\usepackage{amsfonts}
\usepackage{times}
\usepackage{graphicx}
\usepackage{latexsym}
\usepackage{dsfont}
\usepackage{amssymb}
\usepackage{amsmath}
\usepackage{cite}
\usepackage{verbatim}
\usepackage{subfigure}

\def\bb0{{\mathbb{0}}}

\def\bb{{\mathbf{b}}}

\def\b0{{\mathbf{0}}}

\def\sf0{{\mathsf{0}}}

\usepackage{epstopdf}
\usepackage{enumerate}
\usepackage{algorithmicx}
\usepackage{algorithm}
\usepackage{amsmath}
\usepackage[noend]{algpseudocode}
\usepackage{float}
\usepackage{hyperref}
\usepackage{color}
\usepackage{makeidx}
\usepackage{bbm}
\usepackage{graphicx}
\usepackage{booktabs}
\usepackage{subfigure}

\begin{document}
\title{Parkour Spot ID: Feature Matching in Satellite and Street view images using Deep Learning}
\author{João Morais, Kaushal Rathi, Bhuvaneshwar Mohan and Shantanu Rajesh\\ joao, kaushalr, bhuvanm,  shantanurajesh@asu.edu}

\maketitle

\begin{abstract}
How to find places that are not indexed by Google Maps? We propose an intuitive method and framework to locate places based on their distinctive spatial features. The method uses satellite and street view images in machine vision approaches to classify locations. If we can classify locations, we just need to repeat for non-overlapping locations in our area of interest. We assess the proposed system in finding Parkour spots in the campus of Arizona State University. The results are very satisfactory, having found more than 25 new Parkour spots, with a rate of true positives above 60\%. 
\end{abstract}


\section{Introduction} \label{sec:Intro} 

\subsection{Motivation}
Nowadays, the prevalent method of finding information is to "Google it". Should we want to find locations, Google Maps is our go-to place. However, not all locations are indexed in Google Maps. Indeed, why would Google index all street lamps in New York, all the park benches in Paris, or all the bridges in Amsterdam? Nonetheless, it is conceivable that a photographer is looking for lamps that offer his desired color composition and background or a film crew looking for the ideal bridge to perform a stunt. In this paper, we provide a method for finding non-indexed places in Google Maps.

When locations have distinctive spatial features a straightforward approach to classify locations is to use images \cite{PlaNet}. Such images may have a multitude of sources as long as they can be associated with a GPS location \cite{IM2GPS}. Google APIs allow us precisely that. Through Google Static Maps API \cite{GoogleAPIMaps} and Google Street view API \cite{GoogleAPIStreet}, we can gather satellite and street view images, respectively, to aid classification.

We consider the task of finding outdoor places to practice Parkour and FreeRunning (detailed in Section \ref{sec:ProblemFormulation}). Parkour involves jumping, climbing, and/or running, or any other form of movement, typically in urban environments \cite{ParkourPlayingWithFear}. As a community that just now creating its second generation of practitioners, the Parkour community is new and fast-growing \cite{PKwiki}. As it settles down and organizes, tools and common shared resources emerge, and the most commonly shared information between communities is training locations \cite{ParkourPlayingWithFear}. One of the aims of this work is to help members of the Parkour Community systematically find and share training spots within their region of interest.

\subsection{Prior Work}
Geolocation-aided machine vision has been previously studied. Several studies \cite{PlaNet, IM2GPS} were able to place confidence regions on the surface of the earth, based on the pixels of a single image. The authors of \cite{databaseIMAGEgeolocation} further leverage a hierarchical database to improve geolocation. However, the objective of all these approaches is to use an image to find a location. On the other hand, we aim to find a location, based not on an image, but on a generic set of spatial features and have the system return possible locations within a region of interest. 

\subsection{Contribution}

This work adds to the extensive literature on machine vision applications. It presents no novelty in the methods it employs but in how known methods are utilized to help a growing community. More specifically, in this manuscript, we present:
\begin{itemize}
    \item A scalable method for feature matching in Google Maps;
    \item A real-world-tested and systematic approach towards finding Parkour spots in a region of interest.
\end{itemize}

\begin{figure}
	\centering
	\includegraphics[width=.5\columnwidth]{"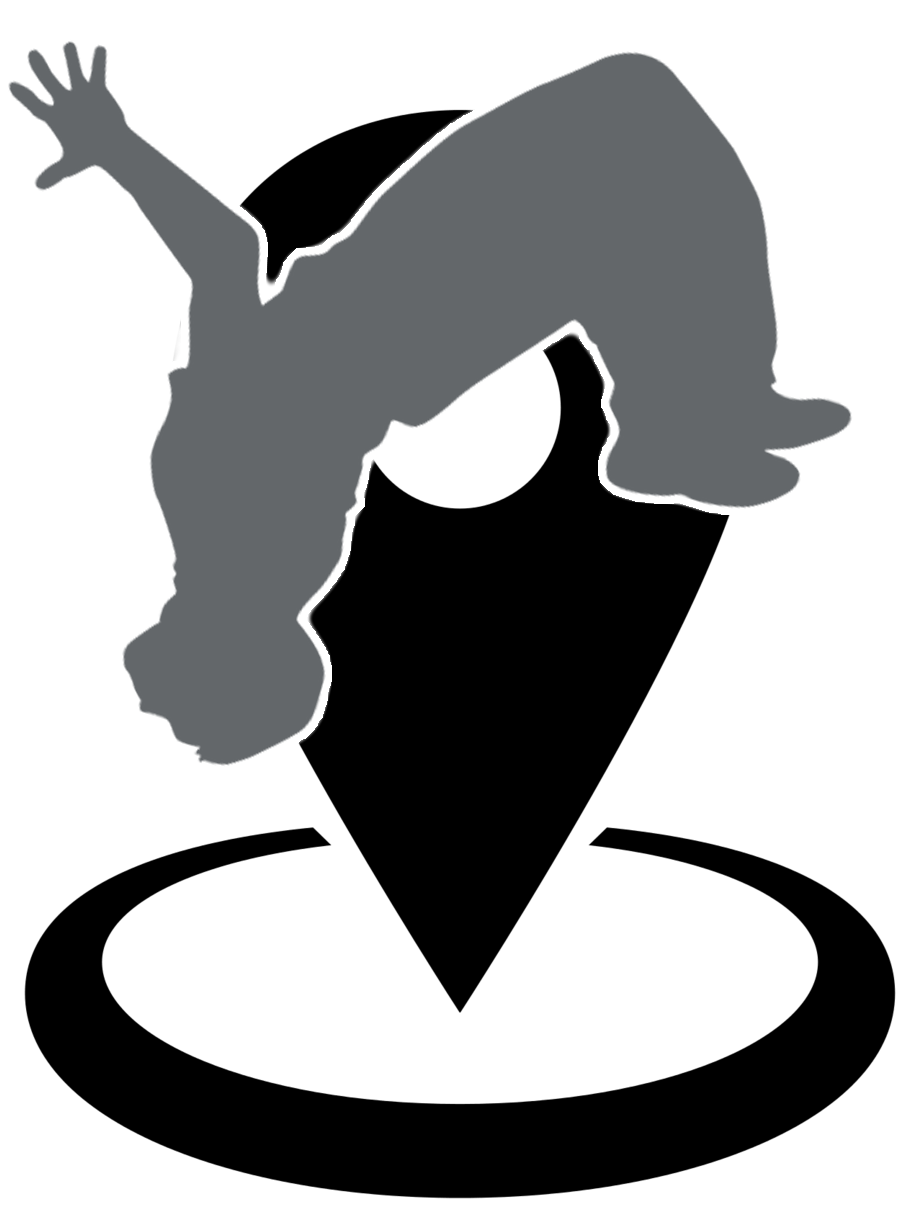"}
	\caption{The Parkour database logo. In GitHub \cite{ParkourSpotIDgithub}.}
	\label{fig:pkdb}
	\vspace{-.5cm}
\end{figure}

Furthermore, we test the system in the Arizona State University (ASU) campus, one of the largest in the United States. And we verify that the method to find Parkour spot presented in this work can methodically populate a database. The method, and the database, have the logo in Figure \ref{fig:pkdb}.

This work is organized as follows. First, in Section \ref{sec:ProblemFormulation} we formulate the problem of geolocation based on feature matching. Section \ref{sec:ProposedSolution} reviews the current literature and presents the proposed solution, detailing the machine learning models employed. In Section \ref{sec:Results} we present the results, both of the individual parts of the system as well as the system as a whole. Lastly, Section \ref{sec:Conclusion} summarizes the conclusions, and in Section \ref{sec:FutureWork} we leave some remarks on how the system can be improved and possible ways of building upon this project.

\section{Problem Formulation} \label{sec:ProblemFormulation}
\subsection{The geo-classification problem}
The problem we aim to solve is a classification task. Given location $\mathbf{x}$ in geographical coordinates (latitude and longitude) such that $\mathbf{x} = (x_{lat}, x_{lon})$, output a probability $\mathbb{P}(\mathbf{x}) \in [0,1]$ that translates the likelihood of $\mathbf{x}$ being a Parkour spot. For the sake of simplicity, we assume the only accessible information about $\mathbf{x}$ are satellite images $I_{sat}$ and street images $I_{str}$. Each image belongs to $\mathbb{R}^3$ - it contains the three values $(r,g,b)$ referring to the components of red, green, and blue (RGB), each between 0 and 1, for every pixel. If we denote by $\mathcal{I}_{sat}$ and $\mathcal{I}_{str}$ the sets of all satellite and street view images of a certain location, then we may further define the functions that will extract knowledge from each set, respectively, $F_{sat}$ and $F_{str}$. Thus, we may write: 

\begin{equation}
    \mathbb{P}(\mathbf{x}) = G\left(F_{sat}(\mathcal{I}_{sat}(\mathbf{x})), F_{str}(\mathcal{I}_{str}(\mathbf{x}))\right),
\end{equation}
where $G: \mathbb{R}^2 \rightarrow \mathbb{R}$ is the function that weighs and combines the outputs of $F_{sat}$ and $F_{str}$ resulting in the probability $\mathbb{P}(\mathbf{x})$.

\subsection{Features of Parkour Spots}
Specific to our application, we must define what features constitute a Parkour spot because the capability of identifying such features needs to be encoded in $F$. Since Parkour, or \textit{l'art du déplacement}, as the first practitioners call it \cite{ParkourTheEvent}, suffers from being quite loosely defined. Despite contributing to its glamour, it complicates objective definitions. Thus, although highly subjective, we attempt to define that the quality of a parkour spot.
\textbf{Definition:} \textit{The suitability of a location for Parkour is proportional to how easily the practitioner can come up with ideas for Parkour moves and sequences in that location.} Therefore, we may objectively conclude that the more architectural features exist to jump, climb, roll, crawl, or interact with the environment, the higher the likelihood of the location to be suitable for Parkour. 

\section{Proposed Solution} \label{sec:ProposedSolution}

Solving the single-coordinate classification problem presented in Section \ref{sec:ProblemFormulation} enables us to check coordinates for Parkour spots. Therefore, to find multiple Parkour spots, one simply needs to run the same algorithm for other coordinates systematically. In this section, we present our proposed solution to the classification of single coordinates. Our solution consists of two computer vision tasks, one for top view and one for street view images, and both tasks rely on object detection methods.

\subsection{Object Detection}

\begin{figure}
	\centering
	\includegraphics[width=1\columnwidth]{"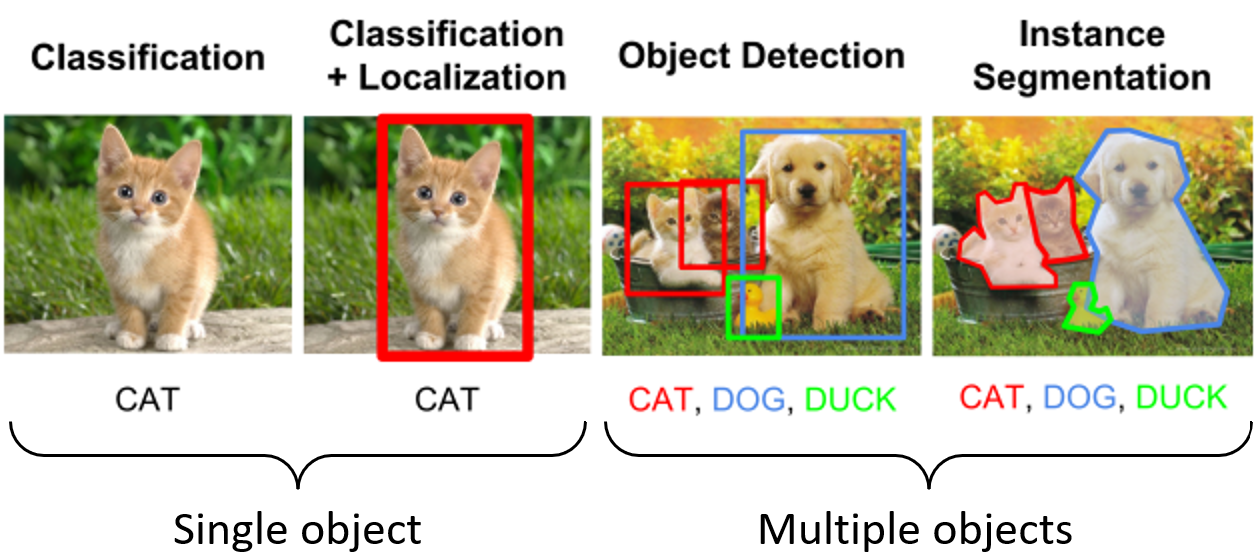"}
	\caption{Differentiation of object detection tasks in computer vision. \cite{objDetection}}
	\label{fig:vision_tasks}
	\vspace{-.5cm}
\end{figure}

Typical object detection tasks have two phases: i) Object Localization and ii) Image Classification \cite{machineVIStasks}. While object localization involves using of a bounding box to locate the exact position of the object in the image, image classification is the process of correctly classifying the object within the bounding box \cite{machineVIStasks}. Figure \ref{fig:vision_tasks} helps show the difference. Instance and semantic segmentation go a level deeper. In semantic segmentation each pixel is classified to a particular class label, hence it is a pixel-level classification \cite{objDetection}. Instance segmentation is similar except that multiple objects of the same class are considered separately as individual entities \cite{objDetection}. 

For the top-view model, we opt an image classification method. For us humans is hard to delineate and annotate useful features for Parkour in satellite images, therefore we intend to leave this complexity for the network to learn. We opt for the instance segmentation approach for the street view model because it provides a more detailed classification, while object detection would only provide bounding boxes. Bounding boxes become less practical and robust when the shape of the object varies considerably and is random in nature \cite{richFeatures}. We choose instance instead of semantic segmentation because the number of objects definitely matters in the quality of Parkour spots.

\subsection{Satellite Imaging model}

The model to process satellite images uses binary classification. In a satellite image, Parkour locations may contain visible features such as stairs, railings, walls, and other elevations that resemble obstacle courses or may be suitable for Parkour. By providing only 0 or 1 labeled image our goal is to have the model identify these patterns through convolutions. 

\begin{figure}
	\centering
	\includegraphics[width=1\columnwidth]{"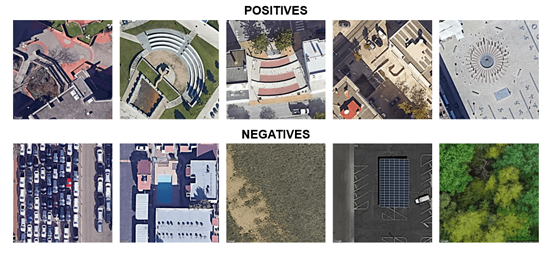"}
	\caption{Positive and negative satellite image examples used for training.}
	\label{fig:pos_neg_examples}
	\vspace{-.4cm}
\end{figure}

The coordinates of known Parkour locations were crowdsourced from parkour communities worldwide. We gathered over 1300 coordinates from cities such as Paris (France), London (United Kingdom), Lisbon (Portugal), and Phoenix (Arizona, United States). For top-view, the coordinates were queried from the Google Map Static API \cite{GoogleAPIMaps} with a fixed magnification of 21, resulting in high definition images of 640 by 640 pixels. For the negative examples required for training of the top-view model, we uniformly sampled cities, gathering 400 random coordinates from 6 random locations, resulting in 2400 negative samples. Figure \ref{fig:pos_neg_examples} shows some positive and negative examples. 

\begin{figure}
	\centering
	\includegraphics[width=1\columnwidth]{"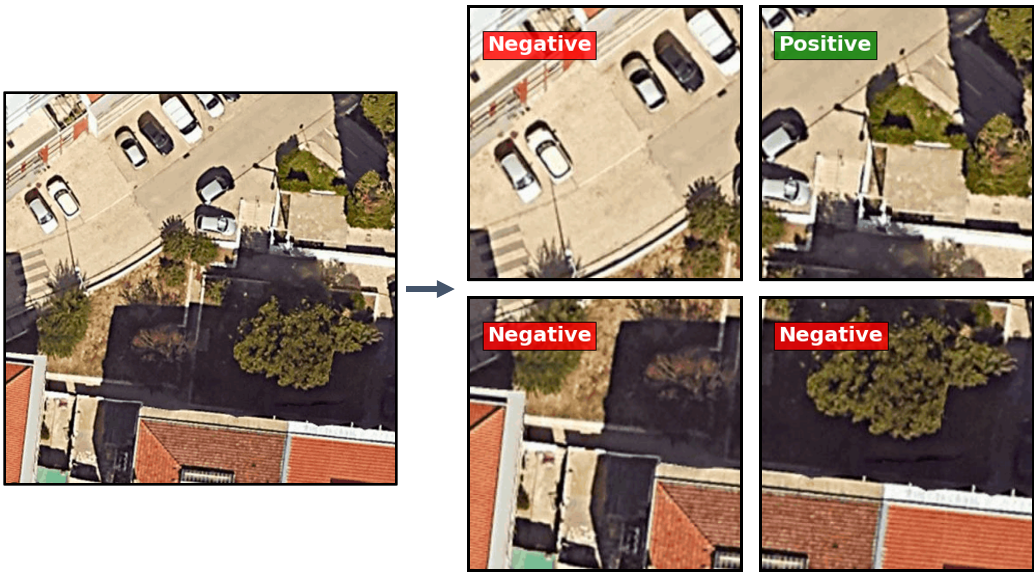"}
	\caption{Splitting a positive sample into quadrants.}
	\label{fig:split_in_four}
	\vspace{-.5cm}
\end{figure}

With our classification problem requiring all the detail satellite images can provide, it is essential to maintain a relatively high resolution. However, larger images imply larger memory requirements during training. We downscaled the images from (640,640) pixels to (512,512). Then, we divide each satellite image into four to reduce the chance of false positives by limiting the information in each input - this approach is represented in Figure \ref{fig:split_in_four}. Post filtering, the positive training set had 3117 samples, and the negative set had 13,231 samples. For training, 3117 random negative samples were selected to maintain class balance. 

During training, multiple models such as VGG16 \cite{vgg16}, Resnet50 \cite{resnet50} and InceptionV3 \cite{inceptionV3}, were tested. We experimented with:
\begin{itemize}
    \item Training partial sections of the model, including (but not limited to) exclusively convolution layers;
    \item Forcing data imbalance towards the positive samples space so the model can learn positive features better;
\end{itemize}
Ultimately, we designed our own model based on the above-mentioned architectures. The hyperparameters used to train the CNN are listed in Table \ref{tab:top-nn}.

\begin{table}[h]
	\caption{Satellite/Top-view model Neural Network Hyper-parameters}
	\centering
	\setlength{\tabcolsep}{5pt}
	\renewcommand{\arraystretch}{1.2}
	\begin{tabular}{@{}l|cc@{}}
		\toprule
		\toprule
		\textbf{Parameters}                     & \textbf{Values}  \\ 
		\midrule \midrule
		\textbf{Input Size}                     & (256,256,3)     \\
		\textbf{Epochs}                         & 100     \\
		\textbf{Batch size}                     & 32 \\
		\textbf{Initial learning rate}          & 0.001    \\
		\textbf{Optimizer}                      & Adam \cite{kingma2017adam}    \\
		
		\bottomrule \bottomrule
	\end{tabular}
	\label{tab:top-nn}
	\vspace{-4mm}
\end{table}

\subsection{Street view model}

One standard method for instance segmentation is using Mask Region-based CNN (R-CNN) \cite{MaskRCNN}. We use an implementation in \cite{MaskRCNNimplementation}. We used transfer learning on a model pre-trained on the COCO dataset \cite{COCOdataset}, and we retrain only the convolutional layers. The COCO dataset consists of 80 distinct categories, but we optimized the model to differentiate only three (short walls, railings, and stairs) from the background (also considered a class).

The data collected from the community had street view images and other images taken by the community. Out of 1300 coordinates verified to contain Parkour spots, each image was manually evaluated, and the dataset was narrowed down to 249 images for training and 51 images for validation. The images were filtered on the basis of:
\begin{itemize}
    \item How clear and understandable the images were to the naked eye.
    \item Selecting only daytime images since nighttime images were really low in number and it could hamper the prediction capability of the model because those images could act as noise. Google street view images are all captured in the daytime.
    \item Images which were blurry and really complicated to annotate were discarded.
\end{itemize}

\begin{figure}
	\centering
	\includegraphics[width=1\columnwidth]{"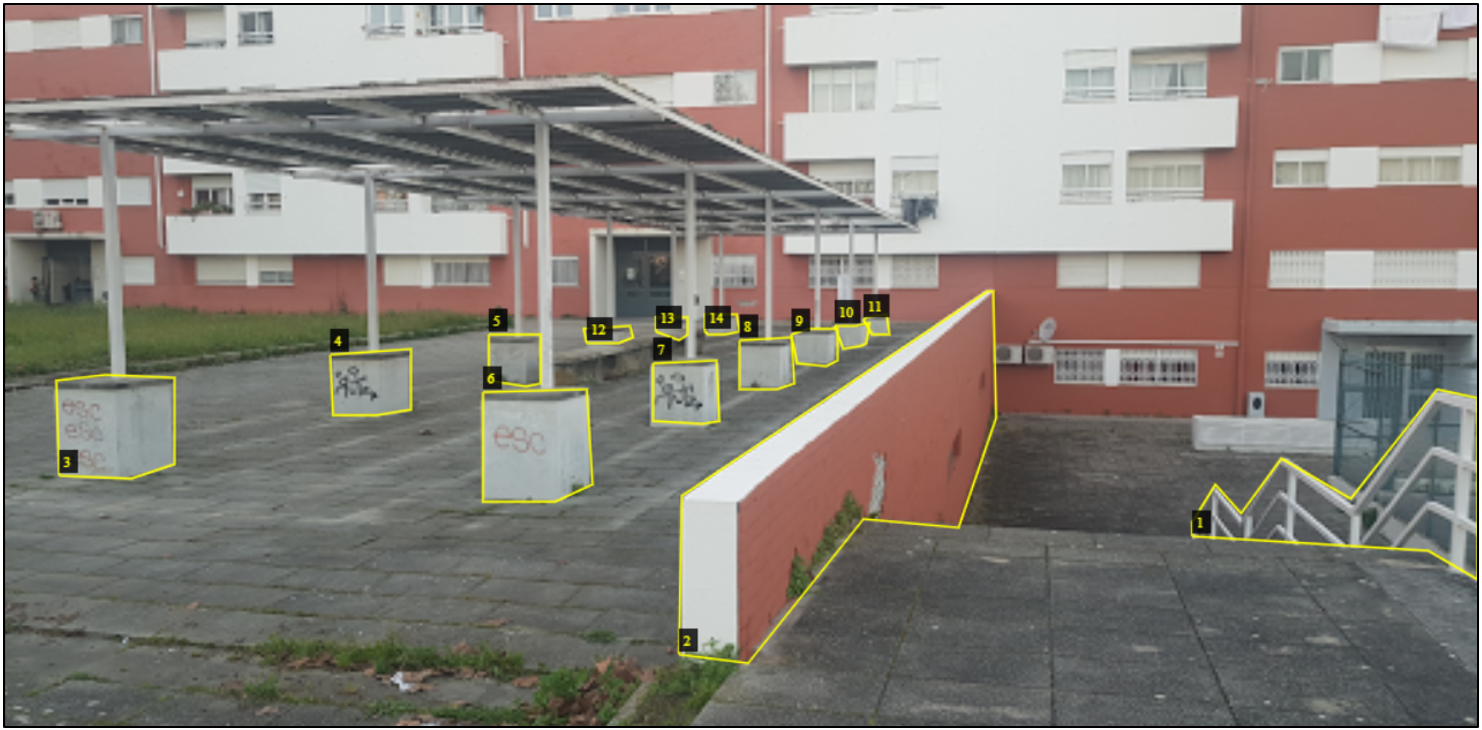"}
	\caption{Example of annotations performed in street view image.}
	\label{fig:street_labeling}
	\vspace{-.5cm}
\end{figure}

After data filtering, the VGG Image Annotator (VIA) \cite{viaANNOTATOR} was used to manually annotate each image. An example of an annotated image is presented in Figure \ref{fig:street_labeling}. 

The Mask R-CNN works for any input up to 1024 by 1024, but we used inputs of size 640 by 640, because that is the maximum size for the Google street view API. 

Contrary to typical model training, we did not aim to minimize the loss during training. Mathematically defining a loss for finding a parkour spot is hard. Instead, we manually tuned parameters and assessed the output images of the test set. When the loss was sufficiently low, and the output matched our intuition for parkour spots by not over- nor under-identifying objects, then we stopped training. Model hyperparameters and implementation-specific parameters are in Table \ref{tab:mrcnn}. For specific parameters, refer to \cite{MaskRCNN, MaskRCNNimplementation} for their meaning. 

\begin{table}[!t]
	\caption{Mask R-CNN Specific parameters and Training Hyper-parameters}
	\centering
	\setlength{\tabcolsep}{5pt}
	\renewcommand{\arraystretch}{1.2}
	\begin{tabular}{@{}l|cc@{}}
		\toprule
		\toprule
		\textbf{Parameters}                     & \textbf{Values}  \\ 
		\midrule \midrule
		\textbf{Input size}                     & (640,640,3)     \\
		\textbf{Epochs}                         & 100     \\
		\textbf{Optimizer}                      & Adam     \\
		\midrule
		\textbf{NUM\_CLASSES}                   & 4\\
		\textbf{STEPS\_PER\_EPOCH}              & 15 \\
		\textbf{VALIDATION\_STEPS}              & 1 \\
		\textbf{BATCH\_SIZE}                    & 1 \\
		\textbf{DETECTION\_MIN\_CONFIDENCE}     & 0.75 \\
		\bottomrule \bottomrule
	\end{tabular}
	\label{tab:mrcnn}
	\vspace{-4mm}
\end{table}

\section{Results} \label{sec:Results}
In this section, we first analyze the performance of each system component, i.e. satellite and street view models. Subsequently, both models are integrated as described in Section \ref{sec:ProposedSolution}. The performance is assessed using real, unlabeled data.

\subsection{Satellite Model}

\begin{figure}
	\centering
	\includegraphics[width=0.8\columnwidth]{"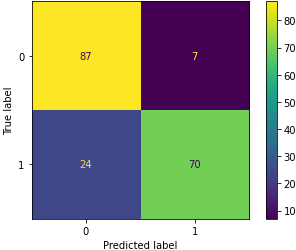"}
	\caption{Confusion matrix of satellite images model.}
	\label{fig:top_conf_mat}
	\vspace{-.5cm}
\end{figure}

The satellite or top-view model was trained on thousands of positive and negative labeled examples. As a result of such training, the binary classification model yielded a classification accuracy of 80\% on our test set. The confusion matrix in \ref{fig:top_conf_mat} reflects the performance of our model.

\begin{figure}
	\centering
	\includegraphics[width=1\columnwidth]{"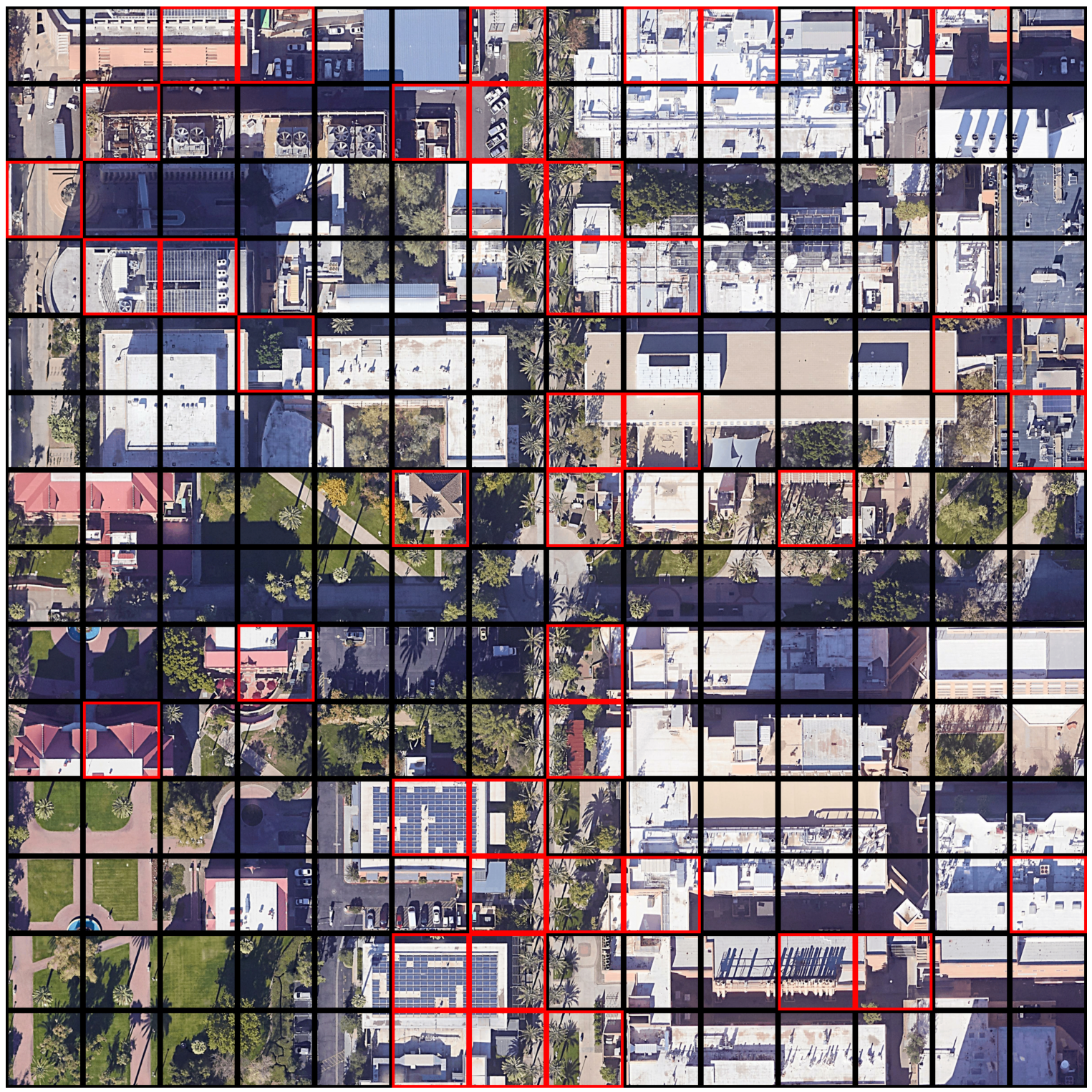"}
	\caption{Results from testing the top-view model.}
	\label{fig:results_top_view}
	\vspace{-.5cm}
\end{figure}

Looking now at unlabeled data, Figure \ref{fig:results_top_view} shows the performance of the model given a grid of 196 images spanning 100 meters from the central coordinate. The model can detect most of the small elevations, walls, railing, stairs, and similar features that are suitable for Parkour. However, the model does classify pointed roofs, solar panel arrays, and HVAC (heating, ventilation and air conditioning) arrays as positives. A solution is to include similar samples in the negatives training set.

\subsection{Street view Model}

\begin{figure}
	\centering
	\includegraphics[width=1\columnwidth]{"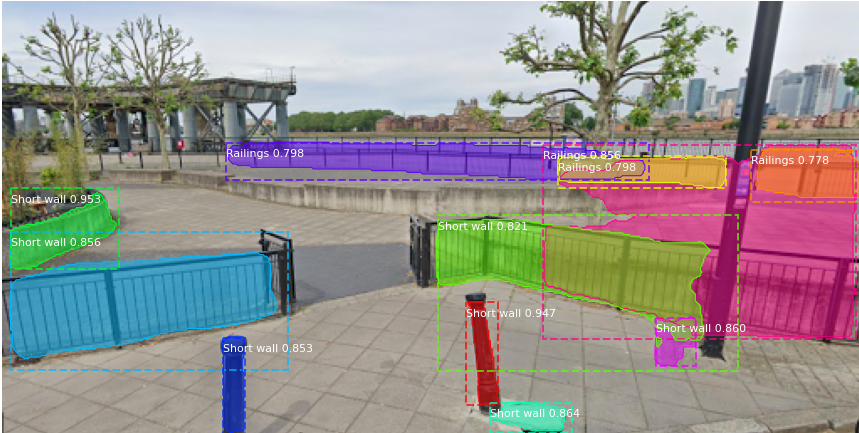"}
	\caption{Example of street view model output in unlabeled data.}
	\label{fig:result_street_view}
	\vspace{-.5cm}
\end{figure}

To assess the street view model in realistic conditions, we used many unlabeled examples. Overall, the model works consistently well. Figure \ref{fig:result_street_view} shows an example. We see that although the model might at times mistake railings by walls, it is still identifying those elements to be useful for Parkour, which is what is most relevant to our application.

\subsection{ASU Campus Results}

To test the end-to-end framework, we used the proposed system to identify spots at ASU campus. We used a center coordinate and an area of interest that is a square inscribed in a circle with a radius of 650 meters. Then we uniformly sampled the region to achieve non-overlapping satellite images (roughly 40 meters apart), and acquired four 90-degree street view images in each of the uniform coordinates. 

The method used to determine the quality of a Parkour spot is counting the number of class hits for each of the four street view directions. If there are more than $T=20$ Parkour-usable objects (i.e. short walls, stairs or rails), we mark the coordinate as containing a Parkour spot. The number of positives can be controlled with the threshold $T$. Resorting solely to street-view provided the highest reliability and interpretability. Table \ref{tab:ASU_stats} shows some statistics from this study.

\begin{table}[!t]
	\caption{Statistics from large-scale testing at ASU campus.}
	\centering
	\setlength{\tabcolsep}{5pt}
	\renewcommand{\arraystretch}{1.2}
	\begin{tabular}{@{}l|cc@{}}
		\toprule
		\toprule
		\textbf{Center coordinate}              & (33.4184, -111.9328)     \\
		\textbf{Radius of interest}             & 650 meters    \\
		\textbf{Number of coordinates}          & 1155     \\
		\textbf{Number of API requests}         & 5775     \\
		\textbf{Total cost of API requests}     & 34.65 \$     \\
		\midrule \midrule
		\textbf{Number of Positives}              & 46     \\
		\textbf{Number of True Positives}         & 28     \\
		\bottomrule \bottomrule
	\end{tabular}
	\label{tab:ASU_stats}
	\vspace{-4mm}
\end{table}

\begin{figure}
	\centering
	\includegraphics[width=1\columnwidth]{"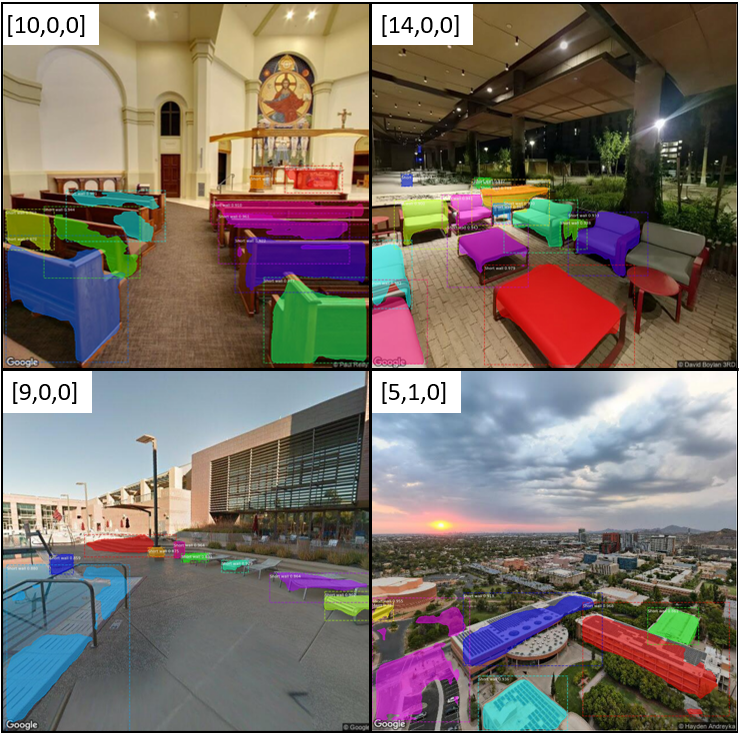"}
	\caption{Examples of false positives in ASU campus test. Label means the number of identified [small walls, rails, stairs].}
	\label{fig:false_positives}
	\vspace{-.5cm}
\end{figure}

Almost 50\% of the positive results are false positives. Figure \ref{fig:false_positives} shows four cases where the system was fooled. First, unbeknownst to us, the Google street view API sometimes returns indoor images. Since tables, benches, counters, and walls are identified as useful for Parkour, indoor locations are wrongly ranked high. Outdoor locations with furniture are also positively classified. Pools too, due to having a fair share of sun loungers and railings. Lastly, several street-view requests had a view considerably above street-level, leading the system to identify spots for 5-story high giants instead of humans. We estimate that filtering problematic inputs from the Google API can reduce the percentage of false positives to below 20\%. 

\section{Conclusions} \label{sec:Conclusion}

In this work, we presented a systematic method for finding Parkour spots. The first of its kind in the Parkour Community. We defined the general feature matching problem. Using a binary classification of satellite images and instance segmentation for street view images and connecting both approaches to maximize the information derived for each coordinate, we accurately determined the likelihood of a location being a Parkour Spot. We then executed our methodology on our campus and personally verified the quality of the results, achieving a precision of over 60\%. Finally, we analyzed the most prominent false positives and identified fixes to improve the system performance further.

\section{Future Work} \label{sec:FutureWork}
We presented a scalable framework because its performance improves by enhancing its parts. One future work direction can be to evolve the proposed system by enhancing the satellite image (top-view) model or the street-view model accuracy, robustness, or inference speed. In terms of inference speed, the bottleneck is on the street-view model. High-speed object detection approaches, like Single Shot Detection \cite{SSD} and YOLO \cite{yolo}, can improve classification speed while possibly improving performance. The integration of both models can be improved to reduce the required API requests, i.e. cost and speed of operation in unknown terrain. Furthermore, another interesting approach is to exploit feature extraction explicitly, e.g. by engineering a solution with edge detection. Finally, it would be interesting to study faster ways of encoding feature knowledge since it takes several days to perform all data annotations.

\bibliographystyle{IEEEtran}
\bibliography{biblio}

\end{document}